\documentclass{article}

\usepackage{arxiv}

\usepackage[utf8]{inputenc} 
\usepackage[T1]{fontenc}    
\usepackage{hyperref}       
\usepackage{url}            
\usepackage{booktabs}       
\usepackage{amsfonts}       
\usepackage{nicefrac}       
\usepackage{microtype}      
\usepackage{lipsum}
\usepackage{graphicx}
\usepackage{graphicx}
\usepackage{comment}
\usepackage{amsmath,amssymb}
\usepackage{color}
\usepackage{multirow,array, bigdelim, makecell, booktabs}
\usepackage{slashbox}
\usepackage{multirow}
\usepackage{booktabs}
\usepackage{rotating}
\usepackage{xcolor,colortbl}
\usepackage{hyperref}
\usepackage{times}
\usepackage{epsfig}
\usepackage{graphicx}
\usepackage{amsmath}
\usepackage{amssymb}
\usepackage[numbers, sort, comma, square]{natbib}

\graphicspath{ {./images/} }

\DeclareRobustCommand{\eg}{\textit{e.g. }}

\DeclareRobustCommand{\ie}{\textit{i.e. }}

\DeclareRobustCommand{\etal}{{et al. }}

\DeclareRobustCommand{\qm}{{QMAR}}
\DeclareRobustCommand{\ntu}{{NTU RGB+D}}
\DeclareRobustCommand{\dmap}{{Densepose}}
\DeclareRobustCommand{\gcolor}{\textcolor[rgb]{0.0,0.1,0.95}}

\title{{Unsupervised View-Invariant Human Posture Representation}}
\date{}

\author{
 Faegheh Sardari \\
  Department of Computer Science\\
  University of Bristol\\
  Bristol, UK \\
  \texttt{faegheh.sardari@bristol.ac.uk} \\
   \And
  Björn Ommer \\
  Heidelberg Collaboratory for Image Processing, IWR\\
  Heidelberg University\\
  Heidelberg, Germany \\
  \texttt{ommer@uni-heidelberg.de} \\
  \And
  Majid Mirmehdi \\
  Department of Computer Science\\
  University of Bristol\\
  Bristol, UK \\
  \texttt{m.mirmehdi@bristol.ac.uk} \\
}

\begin{document}
\date{}
\maketitle
\begin{abstract}
Most recent view-invariant action recognition and performance assessment approaches rely on a large amount of annotated 3D skeleton data to extract view-invariant features. However, acquiring {3D} skeleton data can be cumbersome, if not impractical, in in-the-wild scenarios. To overcome this problem, we present a novel unsupervised approach that learns to extract view-invariant 3D human pose representation from a 2D image without using 3D joint data.
Our model is trained by exploiting the intrinsic view-invariant properties of human pose between simultaneous frames from different viewpoints and their equivariant properties between augmented frames from the same viewpoint. We evaluate the learned view-invariant pose representations for two downstream tasks. We perform comparative experiments that show improvements on the state-of-the-art unsupervised cross-view action classification accuracy on NTU RGB+D by a significant margin, on both RGB and depth images. We also show the efficiency of transferring the learned representations from NTU RGB+D to obtain  the first ever unsupervised cross-view and cross-subject rank correlation results on the multi-view human movement quality dataset, QMAR, and marginally improve on the-state-of-the-art supervised results for this dataset. We also carry out ablation studies to examine the contributions of the different components of our proposed network. 
\end{abstract}
\vspace{-2mm}

\begin{figure}[h]
\begin{center}
    \includegraphics[width=0.85\linewidth]{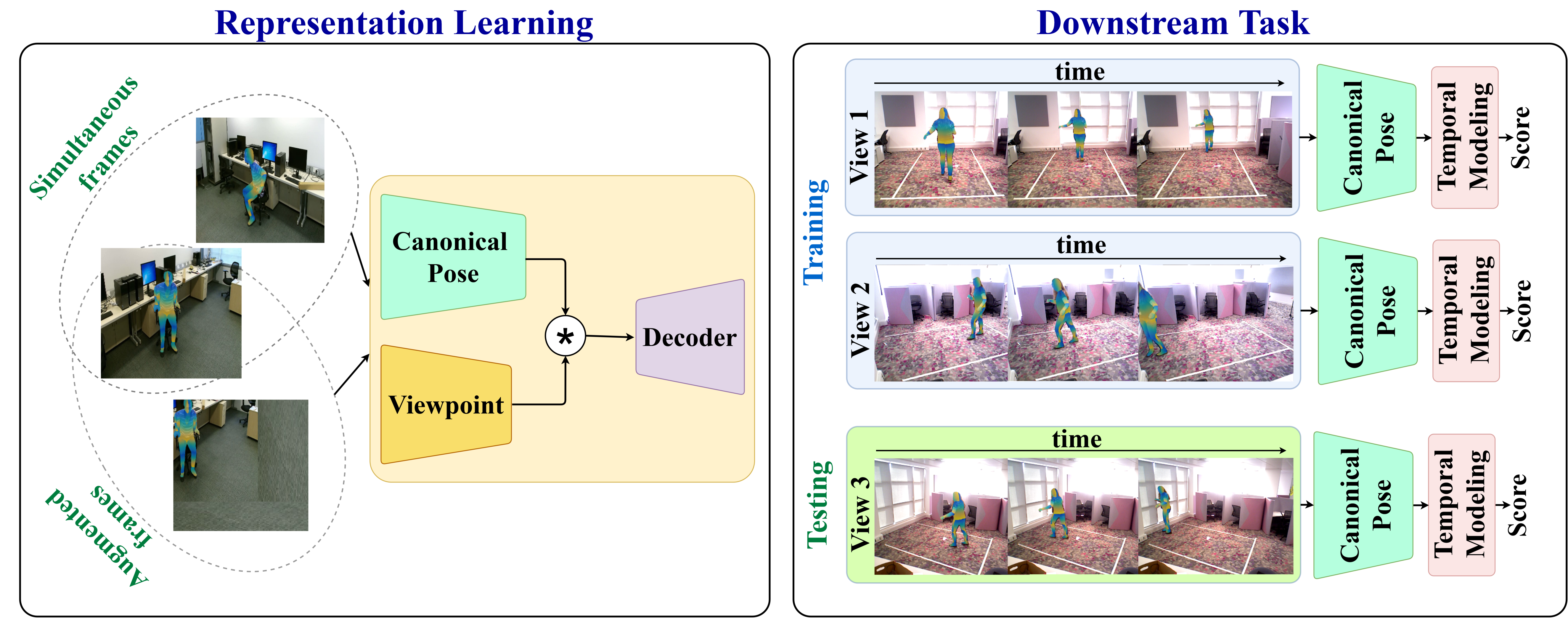}
    \end{center}
    \vspace*{-3mm}
      \caption{{Left: the proposed network learns to disentangle canonical 3D human pose representations and view-dependent features through simultaneous frames from different views and augmented frames from the same view. Right: the unsupervised learned canonical pose representation can be used for downstream tasks.}
      }
    \label{fig:overal figure}
\end{figure}

\vspace{-4mm}

\section{Introduction}
{RGB based deep learning approaches such as \cite{carreira2017quo, feichtenhofer2019slowfast, wu2021towards,kalfaoglu2020late, doughty2019pros, parmar2021piano, parmar2021piano, tang2020uncertainty, pan2019action} have shown an impressive performance in human action recognition and performance assessment. However, as stated in \cite{varol2021synthetic}, the performance of such approaches drops significantly when they are applied on data that come from unseen viewpoints.}
To tackle this problem, a simple solution would be to train a network on data from multiple views \cite{varol2021synthetic}. However, in practice, capturing a labelled dataset of different views is cumbersome and rare - {but two example cases are the commonly used NTU \cite{shahroudy2016ntu} and the recent health-related QMAR \cite{sardari2020vi} datasets, where the granularity of the labelling is still coarse at action class level and at overall performance score level respectively. Ideally, a wholly view-invariant approach would be trained on data from as few views as possible and be able to perform well on a single (unseen) view at inference time. We shall use the NTU and QMAR datasets in our downstream tasks for our proposed view-invariant pose representation method towards this ideal.}
 
Most current view-invariant action classification approaches are based on supervised learning, with both training and testing commonly carried out using skeleton data, such as \cite{zhang2019view, zhang2020semantics, nie2021view,gedamu2021arbitrary, huang2021view}. Others like \cite{rahmani2018learning, sardari2019view, das2020vpn, das2019focus} train on both RGB and 3D joint annotations to facilitate testing using RGB images alone. However, all these approaches rely on a significant amount of annotations (and sometimes camera parameters) during training, the provision of which is expensive and difficult in in-the-wild scenarios. Only very few works, such as \cite{wang2018dividing, sardari2020vi, varol2021synthetic} deal with training from RGB images only. Moreover, the authors are aware of no other unsupervised view-invariant RGB-only  action classification or assessment study, and only of one such work based on RGB and depth \cite{li2018unsupervised}.

In this paper, we propose a representation learning approach to disentangle canonical {(view-invariant)} 3D pose representation and view-dependent features from either an RGB-based 2D {\dmap} human representation map {\it or} a depth {mask} image without using 3D skeleton annotations or camera parameters. We design an auto-encoder comprising two encoders and a decoder. The first is a view-invariant 3D pose encoder that learns 3D canonical pose representations from an input image, and the second is a viewpoint encoder that extracts rotation and translation parameters, such that {when they} are applied on the canonical pose features, it would result in view-dependent 3D pose representation {which are fed into the decoder to reconstruct the input image}. To train the network, we impose geometrical and {positional} order consistency constraints on pose representation features through novel view-invariant and equivariance losses respectively. The view-invariant loss is computed based on the intrinsic view-invariant properties of pose features between simultaneous frames from different viewpoints, while the equivariance loss is computed using the equivariant properties between augmented frames from the same viewpoint. After training, the 3D canonical pose representations can be used for downstream tasks such as view-invariant action classification and human movement analysis. Fig. \ref{fig:overal figure} shows the proposed view-invariant pose representation learning framework and its application on a view-invariant downstream task.

Our key contributions can be summarized as follows: (i) we propose a novel unsupervised method that learns view-invariant 3D pose representation from a 2D image {without using 3D skeleton data and camera parameters}. {Our view-invariant features} can be applied {\it{directly}} by downstream tasks to be resilient to human pose variations in unseen viewpoints, {unlike unsupervised 3D pose estimation methods such as \cite{rhodin2018unsupervised,chen2019unsupervised, chen2019weakly, tripathi2020posenet3d, honari2021unsupervised, dundar2021unsupervised} which obtain view-specific 3D pose features, and require camera parameters and further steps to align  their {view-specific} {features} in a canonical space}, 
(ii) we introduce novel view-invariance and equivariance losses that impose on the network to preserve geometrical and positional order consistency of pose features - these losses can benefit the training process in other pretext tasks that exploit landmark representation, 
(iii) we evaluate the performance of learned pose features on {two downstream tasks that demand view-invariancy} and achieve  state-of-the-art  unsupervised cross-view action recognition accuracy on the {\ntu} standard benchmark  dataset for RGB and depth images at $74.8\%$ and $67.5\%$  respectively, and for the first time we obtain unsupervised cross-view and cross-subject rank correlation results for human movement assessment scores on the {\qm} dataset, while exceeding its supervised state-of-the-art results, (iv) we perform ablation studies to explore the impact of our loss functions on our proposed model. 
 
 
\section{Related Works}
{We now consider {\it the more recent} related works that deal with unsupervised pose representation and view-invariancy,  in particular in relation to our chosen downstream tasks.} 

{\bf Unsupervised Pose Representation -- }
There are several recent examples of RGB-based unsupervised learning approaches to 3D pose estimation, such as  \cite{rhodin2018unsupervised, chen2019unsupervised, chen2019weakly, tripathi2020posenet3d, honari2021unsupervised, dundar2021unsupervised}. {Authors in \cite{chen2019unsupervised, chen2019weakly, tripathi2020posenet3d} extract unsupervised pose features from 2D joints generated from RGB data. For example, Chen \etal \cite{chen2019unsupervised} train a network through a 2D-3D consistency loss, computed after lifting 2D pose to 3D joints and reprojecting 3D onto 2D.} 
{\citet{dundar2021unsupervised} disentangle pose and appearance features from an RGB image by designing a self-supervised auto-encoder {that reconstructs an input image into} foreground and background with the constraint that the appearance features remain consistent temporally while the pose features change.} \citet{honari2021unsupervised} also relies on temporal information and factorizes the pose and appearance features in a contrastive learning manner. {In \cite{ rhodin2018unsupervised}, the authors design a network to encode 3D pose features by predicting a different viewpoint of the input image, but there is no restriction to generate the same pose representation for the simultaneous frames.
All these methods are view-specific and do not generate the same (\ie canonical) 3D pose features for different viewpoints, so they cannot be applied to unseen-view downstream tasks, and camera parameters and extra steps are needed to map their view-specific output into a canonical view. Our proposed method learns view-invariant pose representation from the input image such that it can be applied {\it directly} to unseen-view tasks, such as action recognition.}
{\citet{rhodin2018learning} use both labelled and unlabelled data to estimate canonical 3D pose. They train a network that maps multiple views into a canonical pose through mean square error, but as using only this constraint may generate random features without any positional order consistency, they also use a small subset of 3D pose annotations to enhance the output. However, our proposed approach achieves positional order consistency without utilizing any labels.} {{Note,} positional order consistency {is important} as it enables us to leverage temporal aspects of corresponding body joints to handle video-based downstream tasks.}

{\bf Supervised View-Invariant Action Recognition and Performance Assessment -- } 
To deal with view-invariancy, most action recognition methods {are based on} 3D skeleton joints, such as \cite{rahmani2018learning, zhang2019view, li2019actional, ji2019attention, sardari2019view, zhang2020semantics, huang2021view, gedamu2021arbitrary}. For example, \citet{zhang2019view} present a dual-stream network, one LSTM and one CNN, and fuse the results to predict the action label. Both streams include a view adaptation network estimating the transformation parameters of skeleton data to a canonical view, followed by a classifier. In general, methods that rely on skeleton annotations must rely on fulsome 3D joint representations which are difficult to come by in in-the-wild scenarios. Recently a few works have developed view-invariant action recognition {or analysis} approaches from RGB-D images, such as \cite{varol2021synthetic, wang2018dividing, sardari2020vi, dhiman2020view}. \citet{varol2021synthetic} deploy multi-view synthetic {videos} for training their network to perform action recognition given novel viewpoints, but still use 3D pose annotations to produce the synthetic data, {while the newly generated videos would have to be also labelled by experts if they were to be used for specialist applications such as healthcare.} 

{In view-invariant action performance assessment, we are aware of only one study where \citet{sardari2020vi} investigated a supervised model to assess and score the quality of movement in {subjects simulating Parkinson and Stroke symptoms} by evaluating canonical spatio-temporal trajectories derived from body joint heatmaps.}

{\bf Unsupervised View-Invariant Action Recognition -- }
{There are also only a relatively few  unsupervised deep learning approaches that challenge view-invariant action recognition, e.g. \cite{li2018unsupervised, cheng2021hierarchical}. For instance, Li \etal \cite{li2018unsupervised} introduce an RGB-D based auto-encoder network that extracts unsupervised view-invariant spatio-temporal features from a video sequence. The proposed network is trained to reconstruct two simultaneous source and target view sequences from a given source view video. This method requires both RGB and depth data for training.}
\citet{cheng2021hierarchical} {propose a 3D skeleton-based unsupervised approach by using motion prediction as a pretraining task to learn temporal dependencies for long video representation with a transformer-based auto-encoder.} 


\section{Proposed Method} \label{sec:main}
\begin{figure}[h]
\begin{center}
    \includegraphics[width=0.95\linewidth]{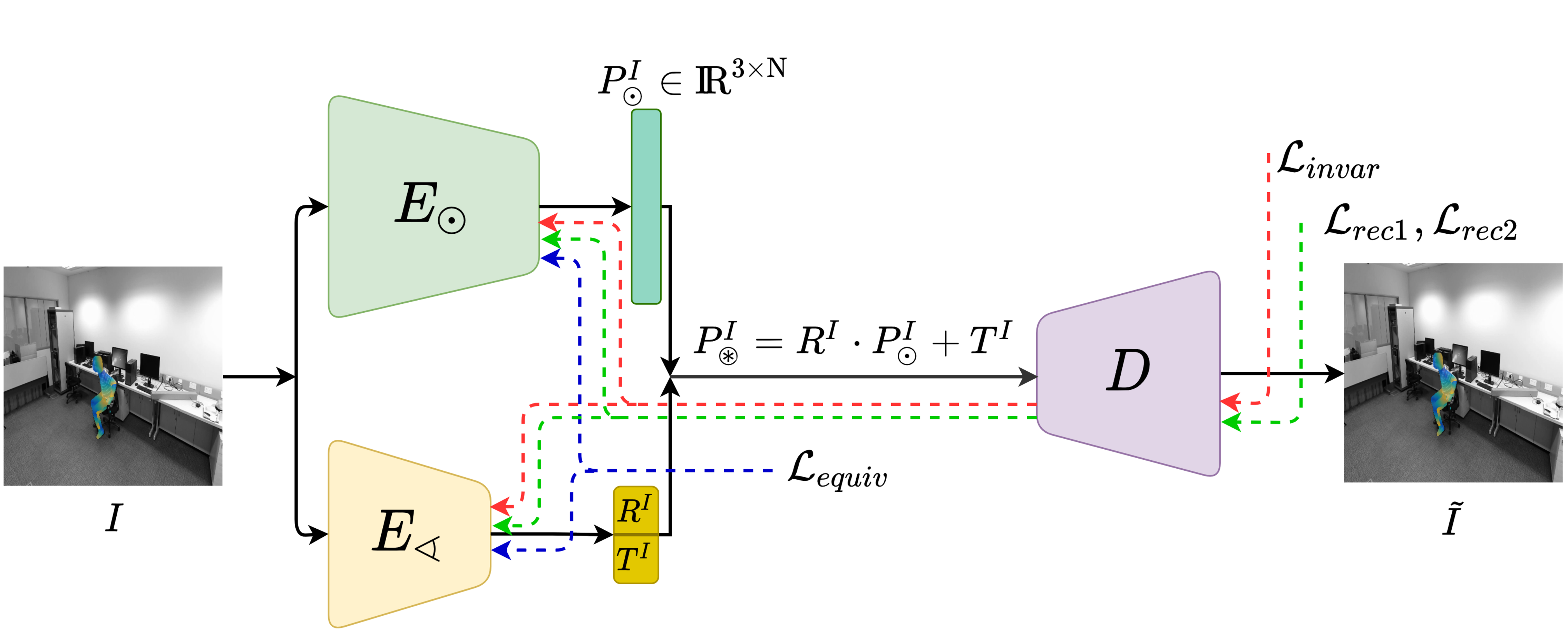}
    \end{center}
    \vspace{-2mm}
      \caption{The overall schema of the proposed view-invariant posture representation learning architecture.}
    \label{fig:autoencoder}
\end{figure}

Our aim is to learn view-invariant 3D pose representation from 2D RGB or depth images without relying on 3D skeleton annotations and camera parameters. {Our method leverages on} geometric transformation amongst different viewpoints and the equivariant property of human pose. The proposed auto-encoder includes a view-invariant pose encoder $E_\odot$, a viewpoint encoder $E_\sphericalangle$, and a decoder $D$ arranged as shown in Fig.~\ref{fig:autoencoder}. $E_\odot$ learns 3D canonical pose features from a given image which can be either an RGB-based 2D {\dmap} human representation map \cite{neverova2019correlated} or a depth {mask} image. As the extracted pose features are canonical, they are mapped into a specific viewpoint using the parameters obtained through encoder $E_\sphericalangle$ before being passed to $D$ to allow the decoder to reconstruct the input image. The network optimises through four losses to generate its view-invariant representation.

\noindent {\bf Model Architecture and Formulation -- } 
 The view-invariant pose encoder $E_\odot$ learns 3D canonical pose features  ${P}^{I}_{\odot} = E_{\odot}(I)$ given image $I\in \rm {I\!R}^{3\times W \times H}$ {where ${P}^{I}_{\odot}\in \rm {I\!R}^{3\times N}$ and} $N$ refers to the number of 3D pose features.  $E_\sphericalangle$ estimates the viewpoint parameters $(R^{I}, T^{I}) = E_{\sphericalangle}(I)$, i.e. rotation  $R^{I} = (\theta_x, \theta_y, \theta_z)$ and translation $T^{I} = (t_x, t_y, t_z)$. These viewpoint parameters are applied on the canonical pose features ${P}^{I}_{\odot}$ to transfer them into a specific viewpoint ${P}^{I}_{\circledast}$, such that ${P}^{I}_{\circledast} = R^{I} \cdot {P}^{I}_{\odot} + T^{I}$ where ${P}^{I}_{\circledast}\in \rm {I\!R}^{3\times N}$.
Then, the decoder reconstructs the input, $\tilde{I} = D({P}^{I}_{\circledast})$. {The network's purpose is therefore that it learns to extract the same canonical 3D pose features for simultaneous frames from different viewpoints while maintaining equivariance for the pose features from {their} augmented frames (shifted in position - see details under Equivariance Loss below) from the same viewpoint.}  We train the proposed network by combining {four} losses,  view-invariant  $\mathcal{L}_{invar}$,  equivariance $\mathcal{L}_{equiv}$, and two reconstruction losses $\mathcal{L}_{rec1}$  and  {$\mathcal{L}_{rec2}$}. \\


\noindent {\bf View-Invariant Loss -- }  We start with two simultaneous frames $(I^{v}_k, I^{w}_k)$ from different views $v$ and $w$ of the same scene from their corresponding video sequences at current frame $k$.  These are passed to encoders $E_\odot$ and $E_{\sphericalangle}$ to extract the {canonical} 3D pose features $P^{I^{\phi}_k}_{\odot} = E_{\odot}(I^{\phi}_k)$ and viewpoint parameters $(R^{\phi}_{k}, T^{\phi}_{k})=E_{\sphericalangle}(I^{\phi}_k)$, for $\phi\in\{v,w\}$. 

{Each frame $k$} has a distinct translation parameter, while the rotation is the same for all the frames of a sequence captured from the same viewpoint. Thus, if we estimate the rotation parameters from two random frames $I^{v}_m$ and $I^{w}_n$ from corresponding sequences and views instead, the network should still retrieve the view-specific pose features. We use this constraint to prevent the model leaking any pose information through $E_\sphericalangle$ and force it to concentrate on only the  viewpoint parameters. Hence, with a probability of 0.5, we {randomly select} the frame to predict the rotation parameters {for the two views,}
\begin{equation}
R^{v} = \begin{cases}
                           R^{v}_{k} & \text{if $r$ is < $0.5$}\\                                           R^{v}_{m} & \text{else}
             \end{cases} \text{~~ and ~~}
R^{w} = \begin{cases}
                           R^{w}_{k} & \text{if $r$ is < $0.5$}\\                                           R^{w}_{n} & \text{else}
             \end{cases}, ~~ r \in U(0, 1),
\end{equation}

\noindent {where $U(0, 1)$ denotes a uniform distribution returning a number between 0 and 1.} As we assume $E_\odot$ encodes the same canonical 3D pose features for $I^{v}_k$ and $I^{w}_k$, then swapping their pose features while their viewpoint features are retained, the network has to still be able to reconstruct them. Thus, the view-invariant loss is obtained by
\begin{gather}
    \label{eq:loss invar 1}
    {{}\tilde{I}^{\prime}}^{v}_k = D({P^{\prime}}^{I^{v}_k}_{\circledast}) \text{~ where ~} {P^{\prime}}^{I^{v}_k}_{\circledast} = R^{v} \cdot P^{I^{w}_k}_{\odot} + T^{v}_{k}, ~~  \\
    \label{eq:loss invar 2}
    {{}\tilde{I}^{\prime}}^{w}_k = D({P^{\prime}}^{I^{w}_k}_{\circledast}) \text{~ where ~} {P^{\prime}}^{I^{w}_k}_{\circledast} = R^{w} \cdot P^{I^{v}_k}_{\odot} + T^{w}_{k}, ~~ \\
    \label{eq:loss invar 3}
    \mathcal{L}_{invar} = \sum_{\phi\in\{v,w\}}\hspace{-0.5em}{MSE(I^{\phi}_{k},~{{}\tilde{I}^{\prime}}^{\phi}_{k})} ~,
\end{gather}
{where $MSE$ indicates the mean square error.}

{However, computing only ${\mathcal{L}}_{invar}$ is not enough to learn the view-invariant pose features, and $E_{\odot}$ still has to reconstruct the simultaneous frames even without swapping their canonical pose features, otherwise the network learns to only assign random latent codes for canonical pose features, so we introduce $\mathcal{L}_{rec1}$ as}
\begin{equation}
    {\mathcal{L}}_{rec1} = \sum_{\phi\in\{v,w\}}\hspace{-0.5em}{MSE(I^{\phi}_{k},~\tilde{I}^{\phi}_{k})} ~,
\end{equation}
 where $\tilde{I}^{\phi}_k = D(P^{I^{\phi}_k}_{\circledast})$ with $P^{I^{\phi}_k}_{\circledast} = R^{\phi} \cdot P^{I^{\phi}_k}_{\odot} + T^{\phi}_{k}$ for $\phi\in\{v,w\}$. \\
 
\noindent {\bf Equivariance Loss -- } The effect of this loss is to help teach the network to preserve {the positional order} of the pose components.
For example, if the $i^{th}$ dimension of the latent variable indicates the right shoulder of a subject, it should be consistent for all the images. 
We assume that the proposed network generates consistent order of pose features{, and $x$ and $y$ axes of view-specific 3D pose space are the same as the $x$ and $y$ directions of the 2D images}, so when  $I^{v}_k$ and $I^{w}_k$ shift by some pixels in the $x$ and $y$ directions, then all components of the view-specific pose {$P^{I^{\phi}_k}_{\circledast}$} would shift similarly. Hence, we propose an equivariance loss computed from augmentations of $I^{v}_k$ and $I^{w}_k$, where the augmented images, ${\Dot{I}}^{v}_k$ and ${\Dot{I}}^{w}_k$, represent positional changes of the human subject in the scene, {for example by $c_1$ and $c_2$ pixels respectively}, i.e. 
\begin{equation}
    \label{eq: equiv}
    \mathcal{L}_{equiv} = \hspace{-1.3em} \sum_{\phi\in\{v,w\}, j\in\{1,2\}}\hspace{-1.6em}{MSE(P^{I^{\phi}_k}_{\circledast} + c_j,~ P^{\Dot{I}^{\phi}_k}_{\circledast})}, 
\end{equation}
where $P^{{\Dot{I}}^{\phi}_k}_{\odot} = E_{\odot}({\Dot{I}}^{\phi}_k)$ and $P^{{\Dot{I}}^{\phi}_k}_{\circledast} = R^{I^{\phi}} \cdot P^{{\Dot{I}}^{\phi}_k}_{\odot} + T^{{\Dot{I}}^{\phi}_k}$ for $\phi\in\{v,w\}$.

$\mathcal{L}_{equiv}$ is computed based on the view-specific pose features while we can also benefit {from} the reconstruction of the augmented frames to improve on the pose representation, so {we introduce ${\mathcal{L}}_{rec2}$ as}.
\begin{equation}
    {\mathcal{L}}_{rec2} = \sum_{\phi\in\{v,w\}}\hspace{-0.5em}{MSE({\Dot{I}}^{\phi}_{k},~\tilde{\Dot{I}}^{\phi}_{k})} ~,
\end{equation}
where  $\tilde{{\Dot{I}}}^{\phi}_k = D(P^{{\Dot{I}}^{\phi}_k}_{\circledast})$. {The total loss is computed as}
\begin{equation}
\label{eq:total loss}
    \mathcal{L}_{total} =  \alpha \cdot {\mathcal{L}_{invar}} + \beta \cdot {\mathcal{L}_{equiv}}  + \gamma \cdot({\mathcal{L}_{rec1} + \mathcal{L}_{rec2}})
\end{equation}
{We determine the weights empirically to be $\alpha = 1.0$, $\beta = 0.001$, and $\gamma = 1.0$.}
{After training the proposed network to learn the 3D canonical pose features, $E_{\odot}$ is used for our example view-invariant downstream tasks. Next, in Section \ref{sec:Experiments}, we outline our proposed method to model temporal aspects of the canonical pose features for each downstream task.}\\


\section{Experiments}
\label{sec:Experiments}
{\bf Datasets -- } {Capturing multi-view datasets requires elaborate set-ups and is inevitably time-consuming and potentially quite expensive. Hence, very few multi-view action classification or movement assessment datasets exist. We present results on two existing datasets, {\ntu} and {\qm}.} 


{\ntu ~\cite{shahroudy2016ntu} is the main benchmark dataset for view-invariant action recognition, including 60 action classes performed by 40 subjects. NTU (for short)  contains 17 different environmental settings captured by three cameras from three viewpoints. 
Two standard protocols are used to evaluate the performance of view-invariant action recognition methods on NTU, cross-view (CV) and cross-subject (CS). In CV, {different views are adopted for training and testing, while in CS different subjects are engaged for training and testing. We followed both protocols by using the same training and testing sets as in \cite{shahroudy2016ntu} for both pretext and  downstream tasks.}}

{QMAR \cite{sardari2020vi} is the only RGB+D multi-view dataset known to the authors for quality of movement assessment. It comprises 38 subjects captured from 6 different views, three frontal and three sides. The subjects were trained by an expert to simulate four Parkinsons and Stroke movement tests: {walking with Parkinson (W-P), walking with Stroke (W-S), sit-to-stand with Parkinson (SS-P), and sit-to-stand with Stroke (SS-S),} 
and the movements were annotated  to determine severity of the abnormality scores. For evaluation, we followed \cite{sardari2020vi, parmar2019and} and obtained Spearman’s rank correlation (SRC) results under CV and CS protocols. For CS,  we used the same training and testing sets as in \cite{sardari2020vi}, and for CV, the data from one frontal and one side view were used for training while the rest of the viewpoints were applied for inference (see Supplementary Materials for details}).

\vspace{2mm}
\noindent {\bf Implementation Details {and Hyper-Parameter Settings} -- }
Our auto-encoder is inspired by the U-Net encoder/decoder \cite{ronneberger2015u,esser2018variational, rhodin2018unsupervised, dorkenwald2020unsupervised}. {The U-Net is a convolutional or spatial latent auto-encoder with skip connections between the encoder and the decoder parts, while we desire a {dense} one \cite{baur2018deep} {without the skip connections  to encode the 3D pose features, so we adapted it for our problem}}. Table \ref{tab:network details} shows details of the proposed network architecture.
We implemented our model in Pytorch and trained it for 20 epochs using Adam \cite{kingma2014adam} with a fix learning rate of 0.0002, and batch size 5. During training, we applied random horizontal flipping for data augmentation. {The depth mask images of NTU used in our experiments contain bounding box of subjects as released by \cite{shahroudy2016ntu}.}
\begin{table}[h]
\centering
\scalebox{0.86}{
    \begin{tabular}{|c|l|}\hline
    \textbf{Module} & \multicolumn{1}{c|}{\textbf{{Layers}}} \\ \hline \hline
    \multirow{3}{*}{{\bf $E_{\odot}$}} & $\{C2(3 \times 3,64),~\gcolor{BN}, ~\gcolor{ReLU}\}\times 2$, $MP(2\times 2)$, $\{C2(3 \times 3,128), ~\gcolor{BN}, ~\gcolor{ReLU}\}\times 2$, $MP(2\times 2)$,\\
    & $\{C2(3 \times 3,256), ~\gcolor{BN}, ~\gcolor{ReLU}\}\times 2$, $MP(2\times 2)$, $\{C2(3 \times 3,512), ~\gcolor{BN}, ~\gcolor{ReLU}\}\times 1$,\\
    &  $\{C2(3 \times 3,512), ~\gcolor{ReLU}\}\times 1$, $\{FC(1024), ~\gcolor{ReLU}\}$, $\{FC(512), ~\gcolor{ReLU}\}$, $\{FC(3\times 70)\}$\\ \Xhline{1.5pt}
    {\multirow{2}{*}{\bf $E_{\sphericalangle}$}} & $\{C2(5 \times 5,128),~\gcolor{BN},~\gcolor{ReLU}\}\times 2$, $MP(7\times 7)$, $\{C2(5 \times 5,256), ~\gcolor{BN},~\gcolor{ReLU}\}\times 2$, \\
    & $\{FC(512),~\gcolor{ReLU}, ~\gcolor{Drp}\}$, $\{FC(6)\}$\\ \Xhline{1.5pt} 
    \multirow{2}{*}{{\bf $D$}} & $\{FC(16 \times 16 \times 512), ~\gcolor{ReLU}, ~\gcolor{Drp}\}$, $\{C2(3 \times 3,256), ~\gcolor{BN},~\gcolor{ReLU}\}\times 2$, $\{CT2(3 \times 3,128), ~\gcolor{BN}, ~\gcolor{ReLU}\}\times 2$, \\
    & $\{CT2(3 \times 3,64),~\gcolor{BN},~\gcolor{ReLU}\}\times 2$, $\{CT2(3 \times 3,3), ~\gcolor{BN}, ~\gcolor{ReLU}\}\times 2$, {$tanh$}\\ \hline
    \end{tabular}
     }
    \vspace*{1mm}
    \caption{The proposed auto-encoder's modules -- All modules are 2D. $C2(d \times d,ch)$:  $d\times d$ convolution filters with $ch$ channels, $CT2$: transposed convolution filters, $MP$: max pooling, $BN$: batch normalization, $FC(O)$: FC layer with $O$ outputs.} 
     \label{tab:network details}
\end{table}


{To select the 3D {canonical} pose feature size ${P}^{I}_{\odot}\in \rm {I\!R}^{3\times N}$, we used cross-validation and evaluated the total loss $\mathcal{L}_{total}$ in Eq. \ref{eq:total loss} for $N$ in the range between 40 and 190 with a step-size of 30. The lower bound was inspired by motion capture systems that use 39 markers, and the upper bound was selected based on \citet{rhodin2018unsupervised} who set their latent code size at $3\times 200$. 
As shown in Table \ref{tab:P parameter}, the average $\mathcal{L}_{total}$ cross-validation results on the NTU dataset for both CV and CS protocols is best when $N=70$, hence our 3D {canonical} pose feature size is set at $3\times 70$.}
\begin{table}[h]
\centering
\scalebox{0.86}{
    \begin{tabular}{|c|c|c|c|c|c|c|c|}\hline
    \multicolumn{2}{|c|}{\backslashbox{\bf $\mathcal{L}_{total}$}{$N$}} & \bf{40} & \bf{70} & \bf{100} & \bf{130} & \bf{160} & \bf{190} \\ \hline \hline
    \multirow{2}{*}{\bf RGB} & {CV} & 0.0088 &  \bf{0.0080} & 0.0084 & 0.0081 & 0.0081 & 0.0082\\ \cline{2-8}
    {} & {CS} & 0.0064 & \bf{0.0061}& \bf{0.0061} & 0.0062& 0.0062& 0.0062\\ \Xhline{1.5pt}
    \multirow{2}{*}{\bf Depth} & {CV} & \bf{0.016} & \bf{0.016} & \bf{0.016} & 0.017 & 0.017 & 0.017\\ \cline{2-8}
    {} & {CS} & \bf{0.015} & \bf{0.015} & 0.016 & 0.016 & 0.016 & 0.016\\ \hline
    \end{tabular}
     }
    \vspace*{1mm}
    \caption{{Optimising ${P}^{I}_{\odot}$ - Average $\mathcal{L}_{total}$ cross-validation results on NTU  for different {canonical} pose size (${3\times N}$).}}
     \label{tab:P parameter}
\end{table}

\vspace{2mm}
\noindent {\bf Action Classification --}  {Our proposed auto-encoder can learn unsupervised 3D pose representations without using any action labels. To encapsulate the temporal element of the action recognition downstream task, we added a two-layer bidirectional {gated recurrent unit} (GRU) followed by one FC layer after our view-invariant pose encoder $E_\odot$, and trained it on fixed-size 16-frame input sequences with the cross-entropy loss function. {Similar to \cite{zolfaghari2017chained}}, we subsampled the sequences such that every sequence was divided into 16 segments and one random frame was selected amongst all frames of each segment}.


{All the representation learning methods on NTU that are compared to ours here can operate on either RGB or depth data for training and inference, except \citet{li2018unsupervised} which requires both RGB and depth for its training stage. Providing like-to-like evaluations against these relevant methods is difficult since for all such techniques their method defines  the nature of their backbone architecture, for example they extract spatio-temporal features while we learn pose representation, \eg \cite{vyas2020multiview} uses 3D CNNs whereas ours is integrally a 2D design. In the case of \cite{li2018unsupervised} which applies a 2D ResNet with added ConvLSTM \cite{shi2015convolutional}, we  provide results with the closest possible backbone, comprising a 2D ResNet and an LSTM.} 


\begin{table}[h]
    \centering
    \scalebox{0.88}{
    \begin{tabular}{|l l|l|c|c c|c c|c c|} \hline
          \multicolumn{2}{|c|}{\multirow{3}{*}{\bf Method}} & \multicolumn{1}{c|}{\multirow{3}{*}{\bf Backbone}} & \multirow{3}{*}{\bf Input} & \multicolumn{4}{c|}{\bf Supervised ($\%$)} & \multicolumn{2}{c|}{\multirow{2}{*}{\bf Unsupervised ($\%$)}}\\
          & & &  &\multicolumn{2}{c}{\bf scratch } & \multicolumn{2}{c|}{\bf fine-tune} & & \\ 
          &{}& & &{\bf CV} & \multicolumn{1}{c}{\bf CS}  &  {\bf CV} & {\bf CS} & {\bf CV} & {\bf CS}\\ \hline\hline
          {Shuffle \& Learn \cite{misra2016shuffle}}&ECCV 2016& {AlexNet}& Depth & {-} & {-} & {-} & {-} & {~~40.9~~} & {46.2}\\ \hline
          {Luo \etal \cite{luo2017unsupervised}}& CVPR 2017& {VGG + ConvLSTM}& Depth & {-} & {-}& {-} & {-} & {53.2} & {61.4}\\ \hline
          {Vyas \etal \cite{vyas2020multiview}} \gcolor{\checkmark}& ECCV 2020 & {3D CNN + LSTM}& Depth & {-} &  {-}& {\underline{78.7}} & {\underline{71.8}} & {-} & {-}\\ \hline
          {Li \etal \cite{li2018unsupervised}} \gcolor{\checkmark}&NeurIPS 2018 & {2D ResNet + ConvLSTM}&  Depth & 37.7& 42.3 & 63.9& 68.1 & {53.9} & {\underline{60.8}}\\ \hline
          {Ours} \gcolor{\checkmark}&& {2D ResNet + LSTM} & Depth & {\underline{60.4}} &  {\underline{63.1}} & {75.5} & 72.7 & {\underline{58.3}} & {58.0} \\ \hline
          {Ours} \gcolor{\checkmark}& &  {2D CNN + GRU} & Depth & {\bf 76.7} & {\bf 75.9}  &  {\bf 82.5} & {\bf 78.8}  & {\bf 67.5} & {\bf 64.7} \\ \Xhline{2pt}
          {Luo \etal \cite{luo2017unsupervised}}& CVPR 2017& {VGG + ConvLSTM} & RGB  & {-} & {-} & {-} & {-} & {-} & {56.0}\\ \hline
          {Vyas \etal \cite{vyas2020multiview} \gcolor{\checkmark}}& ECCV 2020 & {3D CNN + LSTM} & RGB & {-} &  {-}&  {\bf 86.3} & {\bf 82.3} & {-} & {-}\\ \hline
          {Li \etal \cite{li2018unsupervised}} \gcolor{\checkmark} & NeurIPS 2018& {2D ResNet + ConvLSTM} & RGB & 29.2& 36.6 &  49.3 & 55.5 & {40.7} & {48.9}\\ \hline
          {Ours} \gcolor{\checkmark}&& {2D ResNet + LSTM}& RGB & {\underline{66.5}} &{\underline{66.7}}& {78.2} & 73.8 & {\underline {62.1}} & {\underline {63.0}}\\ \hline
          {Ours} \gcolor{\checkmark}& &{2D CNN + GRU} & RGB  & {\bf 77.0} & {\bf 70.3} & {\underline{83.6}} & {\underline{78.1}}  & \bf{74.8} & \bf{68.3}\\ \hline
    \end{tabular}
    }
    \caption{Action classification accuracy on NTU for RGB and depth based representation learning approaches. The {\gcolor{\checkmark}} symbol highlights  view-invariant methods. {Our unsupervised results were obtained after freezing $E_{\odot}$’s parameters during the downstream task while the supervised results were obtained by both fine-tuning $E_{\odot}$ and training it from scratch. The best and the second-best results are in \textbf{Bold} and \underline{underline} respectively.}}
    \label{tab:ntu results}
\end{table}
{Table \ref{tab:ntu results} shows that for the unsupervised scenario, our 2D CNN  and GRU backbone significantly improves the state-of-the-art across CV and CS tests, at $74.8\%$, $64.3\%$ for RGB,  and $67.5\%$, $64.7\%$ for depth data, respectively. The 2D ResNet + LSTM incarnation of our method also exceeds across the board on the state-of-the-art in unsupervised results on NTU, for example achieving $62.1\%$ in almost direct comparison to \cite{li2018unsupervised}'s $40.7\%$ for cross-view RGB inference.}

{For the supervised learning case, we improve on all other works with depth data whether training from scratch or fine-tuning our network with best results at $82.5\%$ and $78.8\%$ on CV and CS protocols respectively, and attain very competitive results using RGB in comparison to the 3D CNN-based \cite{vyas2020multiview}.} 

{In Table \ref{tab:ntu skeleton results}, we report the results of recent state-of-the-art unsupervised pose representation methods that operate on 3D skeleton data. \citet{yao2021recurrent} perform better than our method in CV mode and \citet{cheng2021hierarchical}'s result is marginally better than ours in CS mode. These result vindicate our approach as a viable alternative to skeleton-based methods which are altogether more cantankerous to deal with in real-world applications than  RGB or depth derived data.}
\begin{table}[h]
    \centering
    \scalebox{0.88}{
    \begin{tabular}{|l l|l|c|c c|} \hline
          \multicolumn{2}{|c|}{\multirow{2}{*}{\bf Method}} & \multicolumn{1}{c|}{\multirow{2}{*}{\bf {Backbone}}} & \multicolumn{2}{c|}{{\bf Unsupervised ($\%$)}}\\
          &  \multicolumn{1}{c|}{} & & \multicolumn{1}{c}{\bf CV} & \multicolumn{1}{c|}{\bf CS}\\ \hline\hline
          {\citet{su2020predict} }& CVPR 2020 & {GRU}& {~~~{\underline{76.1}}~~~} & \multicolumn{1}{c|}{50.7}\\ \hline
          {\citet{lin2020ms2l} }& ACM Multimedia 2020& {GRU} & {-} & \multicolumn{1}{c|}{52.5}\\ \hline
          {\citet{yao2021recurrent} }&ICME 2021 & {GRU + GCN}&  {\bf 79.2} & \multicolumn{1}{c|}{54.4}\\ \hline
          {\citet{cheng2021hierarchical}  \gcolor{\checkmark}}& ICME 2021& {Transformer} & {72.8} & \multicolumn{1}{c|}{\bf 69.3}\\ \hline
          {\citet{rao2021augmented} \gcolor{\checkmark}}& Information Sciences 2021& {LSTM}& {64.8} & \multicolumn{1}{c|}{\underline{58.5}}\\ \hline
    \end{tabular}
    }
       \vspace{1mm}
    \caption{State-of-the-art action recognition accuracy results on NTU for skeleton-based representation learning approaches. The {\gcolor{\checkmark}} symbol highlights view-invariant methods. The best and the second-best results are in \textbf{Bold} and \underline{underline} respectively.}
    \label{tab:ntu skeleton results}
\end{table}
\vspace{2mm}
\noindent {\bf Ablation Study -- }
{We ablate our losses to examine their impact on the learning of our pose features. Table \ref{tab:ablation analysis} shows the unsupervised action classification accuracy on NTU as we drop each or both of $\mathcal{L}_{invar}$ and $\mathcal{L}_{equiv}$.}


{Table \ref{tab:ablation analysis} shows that removing $\mathcal{L}_{equiv}$ from the training process, our results for both CV and CS in both RGB and depth deteriorates. This verifies that  positional order consistency is essential in both cases. We also observe that eliminating $\mathcal{L}_{invar}$ causes our method's performance to drop in all cases, except for the cross-subject case with depth as the input modality. The increase in performance in this scenario may be attributed to the removal of the extra geometrical constraints that are imposed on the features by the extra simultaneous frames through the presence of the $\mathcal{L}_{invar}$ computation.}


\begin{table}[h]
    \centering
    \scalebox{1.0}{
    \begin{tabular}{|r|c|c|c|c|} \hline
            \multicolumn{1}{|c|}{\multirow{2}{*}{\bf Ours with}}& \multicolumn{2}{c|}{\bf Depth}  &  \multicolumn{2}{c|}{\bf RGB} \\
            & \multicolumn{1}{c}{\bf CV($\%$)} & \multicolumn{1}{c|}{\bf CS($\%$)} & \multicolumn{1}{c}{\bf CV($\%$) } & \bf CS($\%$) \\ \hline\hline
          { $\mathcal{L}_{rec1} + \mathcal{L}_{rec2}$ } & {35.4} & {32.1}  & 35.6 & 34.1 \\ \hline
          { $\mathcal{L}_{equiv} + \mathcal{L}_{rec1} + \mathcal{L}_{rec2}$} & {\underline{64.1}}  & \bf 65.5  & {69.1} & {\underline{64.9}}\\ \hline
    	  {$\mathcal{L}_{invar} + \mathcal{L}_{rec1} + \mathcal{L}_{rec2}$}  & 59.6 &  52.5 & {\underline{70.3}} & {63.3}\\ \hline
    	  { $\mathcal{L}_{invar} + \mathcal{L}_{equiv} + \mathcal{L}_{rec1} + \mathcal{L}_{rec2}$} & \bf 67.5 & {\underline{64.7}} & \bf 74.8 & \bf 68.3\\ \hline
    \end{tabular}}
    \vspace{1mm}
    \caption{Ablation studies on different combinations of losses used in the unsupervised learning process. The best and the second-best results are in \textbf{Bold} and \underline{underline} respectively.}
    \label{tab:ablation analysis}
\end{table}

\vspace{2mm}
\noindent {\bf Human Movement Analysis --}
{Here, we aim to study the efficiency of the learned representation on NTU for quality of movement scoring on QMAR. As in the action recognition task, we added a two-layer bidirectional GRU followed by one FC layer on top of $E_{\odot}$ to deal with temporal analysis. The size of the FC layer is equal to the maximum score for a movement type. However, for movement quality assessment, we require to analyse every single frame of a sequence, so we cannot apply any subsampling strategies for this task. We followed \cite{sardari2020vi} to divide each video sequence into non-overlapping 16-frame video clips. Our network was trained on a random 16-frame clip through the cross entropy loss function, and for inference, all 16-frame clips of a video sequence were processed.  The score for a  sequence was estimated by averaging the outputs of the last FC layer, as in \cite{sardari2020vi}. Our work offers the first ever unsupervised results on QMAR. \citet{sardari2020vi} who introduced QMAR present the only other supervised view-invariant results on this dataset. We also show the performance of two other architectures taken from \cite{sardari2020vi}.}

{Table \ref{tab:cross-subject qmar results} shows our unsupervised human movement analysis results for CV and CS protocols on QMAR, reaching an average SRC of $0.54$ and $0.58$ respectively. These are broadly already competitive to the supervised results in Table \ref{tab:cross-subject qmar results}, particularly when compared against the supervised, Kinetic-400 \cite{kay2017kinetics} pretrained, deep I3D network. Finally, the supervised version of our method, where we fine-tune our network weights after transfering the learnt weights by NTU training, our performance exceeds \citet{sardari2020vi} on average and achieves $0.72$ and $0.70$ for CV and CS respectively.}

\begin{table}
    \centering
    \scalebox{1.0}{
    \begin{tabular}{ c c|c|c|c|c|c|c|c|}\cline{3-9}
    &\multirow{2}{*}{} &\multicolumn{1}{c|}{\multirow{2}{*}{\bf Method}} & \multirow{2}{*}{\bf Training} & \multicolumn{4}{c|}{{\bf Action} (SRC)} & \multirow{1}{*}{\bf Average}\\ 
    & & &  &  \multicolumn{1}{c}{\textbf{W-P}}& \multicolumn{1}{c}{\textbf{W-S}}& \multicolumn{1}{c}{\textbf{SS-P}}&\textbf{SS-S}& (SRC)\\ \hline \hline
    \multicolumn{1}{|c|}{\multirow{6}{*}{\bf CV}}& {\multirow{5}{*}{Supervised}} &\multicolumn{1}{c|}{ C3D (after \cite{parmar2019and})} & custom-trained & {0.65} & {0.37} & {0.21} & {0.45} & {0.42}\\ \cline{3-9}
    \multicolumn{1}{|c|}{}& &\multicolumn{1}{c|}{I3D \cite{carreira2017quo}}& fine-tune & {0.87} & {0.71}& {0.40} &{\underline{0.63}} & {0.65} \\ \cline{3-9}
    \multicolumn{1}{|c|}{}&  &\multicolumn{1}{c|}{VI-Net \cite{sardari2020vi} \gcolor{\checkmark}} & scratch & {\bf 0.92} & {\bf 0.81} & {\underline{0.46}} & 0.61 & {\underline{0.70}} \\ \cline{3-9}
    \multicolumn{1}{|c|}{}& &\multirow{1}{*}{Ours} & \multicolumn{1}{c|}{scratch} & {0.81} & {0.58} & {0.16} & {0.53} & {0.52} \\ \cline{3-9}
    \multicolumn{1}{|c|}{}& &{Ours \gcolor{\checkmark}} & \multicolumn{1}{c|}{fine-tune } & {\underline{0.90}}  & {\underline{0.76}} & {\bf 0.58} & {\bf 0.64} & {\bf 0.72}\\ \cline{2-9}
    \multicolumn{1}{|c|}{} & {Unsupervised}&\multicolumn{1}{c|}{{Ours} \gcolor{\checkmark}} & {-}& { 0.78} & { 0.66} & { 0.29} & { 0.54} & { 0.54}  \\ \hline \hline
    \multicolumn{1}{|c|}{\multirow{6}{*}{\bf CS}}& {\multirow{5}{*}{Supervised}} &{ C3D (after \cite{parmar2019and})}& custom-trained & {0.50} & {0.37} & {0.25} & {0.54} & {0.41}\\ \cline{3-9}
    \multicolumn{1}{|c|}{}&  &\multicolumn{1}{c|}{I3D \cite{carreira2017quo}}& fine-tune & {0.79} &{0.47}& {0.54} &{0.55} & {0.58} \\ \cline{3-9}
    \multicolumn{1}{|c|}{} &  &\multicolumn{1}{c|}{VI-Net \cite{sardari2020vi} \gcolor{\checkmark}}& scratch &{\underline{0.87}} &{\underline{0.52}} & {\underline{0.58}} &{\underline{0.69}}& {\underline{0.66}}  \\ \cline{3-9}
     \multicolumn{1}{|c|}{}& & {Ours} & \multicolumn{1}{c|}{scratch} & {0.81} & {0.51} & {0.39} & {0.72} & {0.60} \\ \cline{3-9}
    \multicolumn{1}{|c|}{} & & {Ours \gcolor{\checkmark}} & \multicolumn{1}{c|}{fine-tune}  &{\bf 0.89} & {\bf 0.54} & {\bf 0.62} & {\bf 0.76} & {\bf 0.70} \\ \cline{2-9}
    \multicolumn{1}{|c|}{} & {Unsupervised}&\multicolumn{1}{c|}{{Ours} \gcolor{\checkmark}}& {-} & {0.70} & {0.50} & {0.48} & {0.66} & {0.58}\\ \hline 
     \end{tabular}}
     \vspace{1mm}
    \caption{ Spearman's rank correlation (SRC) between predicted scores and ground truth labels for cross-subject analysis on different actions of QMAR dataset. I3D was pretrained on Kinetic-400 \cite{kay2017kinetics}, and the  {\gcolor{\checkmark}} symbol highlights view-invariant methods. The best and the second-best results are in \textbf{Bold} and \underline{underline} respectively.}
     \label{tab:cross-subject qmar results}
\end{table}

\section{Conclusion}
{Most current {\it view-invariant} action recognition and performance assessment approaches are based on supervised learning and rely on a large number of 3D skeleton annotations. In this paper, we dealt with these through an unsupervised method to learn view-invariant 3D pose representation from a 2D image. Our experiments show that not only can our learned pose representations be applied on unseen view videos from the same training data, but it can also be used in different domains. Our unsupervised approach is particularly helpful in applications where the use of multi-view data is essential and capturing 3D skeletons is challenging, e.g. in healthcare rehabilitation monitoring at home or in the clinic.} 

{In the pretext stage of our model, we require synchronised  multi-view frames to learn {view-invariant} 3D pose representations. For future work, we will investigate extracting view-invariant pose features from a single view or non-synchronized frames to allow learning to become a simpler process for application to any suitable dataset.}

\bibliographystyle{unsrtnat}
\bibliography{paper}

\begin{thebibliography}{50}
\providecommand{\natexlab}[1]{#1}
\providecommand{\url}[1]{\texttt{#1}}
\expandafter\ifx\csname urlstyle\endcsname\relax
  \providecommand{\doi}[1]{doi: #1}\else
  \providecommand{\doi}{doi: \begingroup \urlstyle{rm}\Url}\fi

\bibitem[Carreira and Zisserman(2017)]{carreira2017quo}
Joao Carreira and Andrew Zisserman.
\newblock {Quo Vadis, Action Recognition? a new model and the kinetics
  dataset}.
\newblock In \emph{proceedings of the IEEE Conference on Computer Vision and
  Pattern Recognition}, pages 6299--6308, 2017.

\bibitem[Feichtenhofer et~al.(2019)Feichtenhofer, Fan, Malik, and
  He]{feichtenhofer2019slowfast}
Christoph Feichtenhofer, Haoqi Fan, Jitendra Malik, and Kaiming He.
\newblock Slowfast networks for video recognition.
\newblock In \emph{Proceedings of the IEEE/CVF International Conference on
  Computer Vision}, pages 6202--6211, 2019.

\bibitem[Wu and Krahenbuhl(2021)]{wu2021towards}
Chao-Yuan Wu and Philipp Krahenbuhl.
\newblock {Towards Long-Form Video Understanding}.
\newblock In \emph{Proceedings of the IEEE Conference on Computer Vision and
  Pattern Recognition (CVPR)}, pages 1884--1894, 2021.

\bibitem[Kalfaoglu et~al.(2020)Kalfaoglu, Kalkan, and
  Alatan]{kalfaoglu2020late}
M~Esat Kalfaoglu, Sinan Kalkan, and A~Aydin Alatan.
\newblock {Late Temporal Modeling in 3D CNN Architectures with Bert for Action
  Recognition}.
\newblock In \emph{European Conference on Computer Vision}, pages 731--747.
  Springer, 2020.

\bibitem[Doughty et~al.(2019)Doughty, Mayol-Cuevas, and Damen]{doughty2019pros}
Hazel Doughty, Walterio Mayol-Cuevas, and Dima Damen.
\newblock {The Pros and Cons: Rank-Aware Temporal Attention for Skill
  Determination in Long Videos}.
\newblock In \emph{Proceedings of the IEEE Conference on Computer Vision and
  Pattern Recognition(CVPR)}, pages 7862--7871, 2019.

\bibitem[Parmar et~al.(2021)Parmar, Reddy, and Morris]{parmar2021piano}
Paritosh Parmar, Jaiden Reddy, and Brendan Morris.
\newblock {Piano Skills Assessment}.
\newblock \emph{arXiv preprint arXiv:2101.04884}, 2021.

\bibitem[Tang et~al.(2020)Tang, Ni, Zhou, Zhang, Lu, Wu, and
  Zhou]{tang2020uncertainty}
Yansong Tang, Zanlin Ni, Jiahuan Zhou, Danyang Zhang, Jiwen Lu, Ying Wu, and
  Jie Zhou.
\newblock {Uncertainty-Aware Score Distribution Learning for Action Quality
  Assessment}.
\newblock In \emph{Proceedings of the IEEE Conference on Computer Vision and
  Pattern Recognition}, pages 9839--9848, 2020.

\bibitem[Pan et~al.(2019)Pan, Gao, and Zheng]{pan2019action}
Jia-Hui Pan, Jibin Gao, and Wei-Shi Zheng.
\newblock {Action Assessment by Joint Relation Graphs}.
\newblock In \emph{Proceedings of the IEEE International Conference on Computer
  Vision (CVPR)}, pages 6331--6340, 2019.

\bibitem[Varol et~al.(2021)Varol, Laptev, Schmid, and
  Zisserman]{varol2021synthetic}
G{\"u}l Varol, Ivan Laptev, Cordelia Schmid, and Andrew Zisserman.
\newblock {Synthetic Humans for Action Recognition from Unseen Viewpoints}.
\newblock \emph{International Journal of Computer Vision (IJCV)}, pages 1--24,
  2021.

\bibitem[Shahroudy et~al.(2016)Shahroudy, Liu, Ng, and Wang]{shahroudy2016ntu}
Amir Shahroudy, Jun Liu, Tian-Tsong Ng, and Gang Wang.
\newblock {NTU RGB+D: A Large Scale Dataset for 3D Human Activity Analysis}.
\newblock In \emph{Proceedings of the IEEE Conference on Computer Vision and
  Pattern Recognition (CVPR)}, pages 1010--1019, 2016.

\bibitem[Sardari et~al.(2020)Sardari, Paiement, Hannuna, and
  Mirmehdi]{sardari2020vi}
Faegheh Sardari, Adeline Paiement, Sion Hannuna, and Majid Mirmehdi.
\newblock {VI-Net—View-Invariant Quality of Human Movement Assessment}.
\newblock \emph{Sensors}, 20\penalty0 (18):\penalty0 5258, 2020.

\bibitem[Zhang et~al.(2019)Zhang, Lan, Xing, Zeng, Xue, and
  Zheng]{zhang2019view}
Pengfei Zhang, Cuiling Lan, Junliang Xing, Wenjun Zeng, Jianru Xue, and Nanning
  Zheng.
\newblock {View Adaptive Neural Networks for High Performance Skeleton-Based
  Human Action Recognition}.
\newblock \emph{IEEE Transactions on Pattern Analysis and Machine Intelligence
  (TPAMI)}, 41\penalty0 (8):\penalty0 1963--1978, 2019.

\bibitem[Zhang et~al.(2020)Zhang, Lan, Zeng, Xing, Xue, and
  Zheng]{zhang2020semantics}
Pengfei Zhang, Cuiling Lan, Wenjun Zeng, Junliang Xing, Jianru Xue, and Nanning
  Zheng.
\newblock {Semantics-Guided Neural Networks for Efficient Skeleton-based Human
  Action Recognition}.
\newblock In \emph{{Proceedings of the IEEE Conference on Computer Vision and
  Pattern Recognition}}, pages 1112--1121, 2020.

\bibitem[Nie and Liu(2021)]{nie2021view}
Qiang Nie and Yunhui Liu.
\newblock {View Transfer on Human Skeleton Pose: Automatically Disentangle the
  View-Variant and View-Invariant Information for Pose Representation
  Learning}.
\newblock \emph{International Journal of Computer Vision (IJCV)}, 129\penalty0
  (1):\penalty0 1--22, 2021.

\bibitem[Gedamu et~al.(2021)Gedamu, Ji, Yang, Gao, and
  Shen]{gedamu2021arbitrary}
Kumie Gedamu, Yanli Ji, Yang Yang, LingLing Gao, and Heng~Tao Shen.
\newblock {Arbitrary-View Human Action Recognition via Novel-View Action
  Generation}.
\newblock \emph{Pattern Recognition}, page 108043, 2021.

\bibitem[Huang et~al.(2021)Huang, Zhou, Qin, et~al.]{huang2021view}
Qingqing Huang, Fengyu Zhou, Runze Qin, et~al.
\newblock {View Transform Graph Attention Recurrent Networks for Skeleton-based
  Action Recognition}.
\newblock \emph{Signal, Image and Video Processing}, 15\penalty0 (3):\penalty0
  599--606, 2021.

\bibitem[Rahmani et~al.(2018)Rahmani, Mian, and Shah]{rahmani2018learning}
Hossein Rahmani, Ajmal Mian, and Mubarak Shah.
\newblock {Learning a Deep Model for Human Action Recognition from Novel
  Viewpoints}.
\newblock \emph{IEEE Transactions on Pattern Analysis and Machine Intelligence
  (TPAMI)}, 40\penalty0 (3):\penalty0 667--681, 2018.

\bibitem[Sardari et~al.(2019)Sardari, Paiement, and Mirmehdi]{sardari2019view}
Faegheh Sardari, Adeline Paiement, and Majid Mirmehdi.
\newblock {View-Invariant Pose Analysis for Human Movement Assessment from RGB
  Data}.
\newblock In \emph{International Conference on Image Analysis and Processing},
  pages 237--248. Springer, 2019.

\bibitem[Das et~al.(2020)Das, Sharma, Dai, Bremond, and Thonnat]{das2020vpn}
Srijan Das, Saurav Sharma, Rui Dai, Francois Bremond, and Monique Thonnat.
\newblock {VPN: Learning Video-Pose Embedding for Activities of Daily Living}.
\newblock In \emph{European Conference on Computer Vision (ECCV)}, pages
  72--90. Springer, 2020.

\bibitem[Das et~al.(2019)Das, Chaudhary, Bremond, and Thonnat]{das2019focus}
Srijan Das, Arpit Chaudhary, Francois Bremond, and Monique Thonnat.
\newblock {Where to Focus on for Human Action Recognition?}
\newblock In \emph{IEEE Winter Conference on Applications of Computer Vision
  (WACV)}, pages 71--80. IEEE, 2019.

\bibitem[Wang et~al.(2018)Wang, Ouyang, Li, and Xu]{wang2018dividing}
Dongang Wang, Wanli Ouyang, Wen Li, and Dong Xu.
\newblock {Dividing and Aggregating Network for Multi-View Action Recognition}.
\newblock In \emph{Proceedings of the European Conference on Computer Vision
  (ECCV)}, pages 451--467, 2018.

\bibitem[Li et~al.(2018)Li, Wong, Zhao, and Kankanhalli]{li2018unsupervised}
Junnan Li, Yongkang Wong, Qi~Zhao, and Mohan Kankanhalli.
\newblock {Unsupervised Learning of View-Invariant Action Representations}.
\newblock In \emph{Proceedings of the Advances in Neural Information Processing
  Systems (NeurIPS)}, pages 1254--1264, 2018.

\bibitem[Rhodin et~al.(2018{\natexlab{a}})Rhodin, Salzmann, and
  Fua]{rhodin2018unsupervised}
Helge Rhodin, Mathieu Salzmann, and Pascal Fua.
\newblock {Unsupervised Geometry-Aware Representation for 3D Human Pose
  Estimation}.
\newblock In \emph{{Proceedings of the European Conference on Computer Vision
  (ECCV)}}, pages 750--767, 2018{\natexlab{a}}.

\bibitem[Chen et~al.(2019{\natexlab{a}})Chen, Tyagi, Agrawal, Drover, Stojanov,
  and Rehg]{chen2019unsupervised}
Ching-Hang Chen, Ambrish Tyagi, Amit Agrawal, Dylan Drover, Stefan Stojanov,
  and James Rehg.
\newblock {Unsupervised 3D Pose Estimation With Geometric Self-Supervision}.
\newblock In \emph{Proceedings of the IEEE Conference on Computer Vision and
  Pattern Recognition (CVPR)}, pages 5714--5724, 2019{\natexlab{a}}.

\bibitem[Chen et~al.(2019{\natexlab{b}})Chen, Lin, Liu, Qian, and
  Lin]{chen2019weakly}
Xipeng Chen, Kwan-Yee Lin, Wentao Liu, Chen Qian, and Liang Lin.
\newblock {Weakly-Supervised Discovery of Geometry-Aware Representation for 3D
  Human Pose Estimation}.
\newblock In \emph{Proceedings of the IEEE Conference on Computer Vision and
  Pattern Recognition (CVPR)}, pages 10895--10904, 2019{\natexlab{b}}.

\bibitem[Tripathi et~al.(2020)Tripathi, Ranade, Tyagi, and
  Agrawal]{tripathi2020posenet3d}
Shashank Tripathi, Siddhant Ranade, Ambrish Tyagi, and Amit Agrawal.
\newblock {PoseNet3D: Learning Temporally Consistent 3D Human Pose via
  Knowledge Distillation}.
\newblock In \emph{{International Conference on 3D Vision (3DV)}}, pages
  311--321. IEEE, 2020.

\bibitem[Honari et~al.(2021)Honari, Constantin, Rhodin, Salzmann, and
  Fua]{honari2021unsupervised}
Sina Honari, Victor Constantin, Helge Rhodin, Mathieu Salzmann, and Pascal Fua.
\newblock {Unsupervised Learning on Monocular Videos for 3D Human Pose
  Estimation}.
\newblock \emph{arXiv preprint arXiv:2012.01511}, 2021.

\bibitem[Dundar et~al.(2021)Dundar, Shih, Garg, Pottorff, Tao, and
  Catanzaro]{dundar2021unsupervised}
Aysegul Dundar, Kevin Shih, Animesh Garg, Robert Pottorff, Andrew Tao, and
  Bryan Catanzaro.
\newblock {Unsupervised Disentanglement of Pose, Appearance and Background from
  Images and Videos}.
\newblock \emph{IEEE Transactions on Pattern Analysis and Machine Intelligence
  (TPAMI)}, 2021.

\bibitem[Rhodin et~al.(2018{\natexlab{b}})Rhodin, Sp{\"o}rri, Katircioglu,
  Constantin, Meyer, M{\"u}ller, Salzmann, and Fua]{rhodin2018learning}
Helge Rhodin, J{\"o}rg Sp{\"o}rri, Isinsu Katircioglu, Victor Constantin,
  Fr{\'e}d{\'e}ric Meyer, Erich M{\"u}ller, Mathieu Salzmann, and Pascal Fua.
\newblock Learning monocular 3d human pose estimation from multi-view images.
\newblock In \emph{Proceedings of the IEEE Conference on Computer Vision and
  Pattern Recognition}, pages 8437--8446, 2018{\natexlab{b}}.

\bibitem[Li et~al.(2019)Li, Chen, Chen, Zhang, Wang, and Tian]{li2019actional}
Maosen Li, Siheng Chen, Xu~Chen, Ya~Zhang, Yanfeng Wang, and Qi~Tian.
\newblock {Actional-Structural Graph Convolutional Networks for Skeleton-based
  Action Recognition}.
\newblock In \emph{Proceedings of the IEEE Conference on Computer Vision and
  Pattern Recognition (CVPR)}, pages 3595--3603, 2019.

\bibitem[Ji et~al.(2019)Ji, Xu, Yang, Xie, Shen, and Harada]{ji2019attention}
Yanli Ji, Feixiang Xu, Yang Yang, Ning Xie, Heng~Tao Shen, and Tatsuya Harada.
\newblock {Attention Transfer (ANT) Network for View-Invariant Action
  Recognition}.
\newblock In \emph{Proceedings of the 27th ACM International Conference on
  Multimedia}, pages 574--582, 2019.

\bibitem[Dhiman and Vishwakarma(2020)]{dhiman2020view}
Chhavi Dhiman and Dinesh~Kumar Vishwakarma.
\newblock {View-Invariant Deep Architecture for Human Action Recognition Using
  Two-stream Motion and Shape Temporal Dynamics}.
\newblock \emph{{IEEE Transactions on Image Processing}}, 29:\penalty0
  3835--3844, 2020.

\bibitem[Cheng et~al.(2021)Cheng, Chen, Chen, Wei, Zhang, and
  Lin]{cheng2021hierarchical}
Yi-Bin Cheng, Xipeng Chen, Junhong Chen, Pengxu Wei, Dongyu Zhang, and Liang
  Lin.
\newblock {Hierarchical Transformer: Unsupervised Representation Learning for
  Skeleton-Based Human Action Recognition}.
\newblock In \emph{IEEE International Conference on Multimedia and Expo
  (ICME)}, pages 1--6. IEEE, 2021.

\bibitem[Neverova et~al.(2019)Neverova, Novotny, and
  Vedaldi]{neverova2019correlated}
Natalia Neverova, David Novotny, and Andrea Vedaldi.
\newblock {Correlated Uncertainty for Learning Dense Correspondences from Noisy
  labels}.
\newblock \emph{{Proceedings of the Advances in Neural Information Processing
  Systems (NeurIPS)}}, 2019.

\bibitem[Parmar and Morris(2019)]{parmar2019and}
Paritosh Parmar and Brendan~Tran Morris.
\newblock {What and How Well You Performed? A Multitask Learning Approach to
  Action Quality Assessment}.
\newblock In \emph{Proceedings of the IEEE Conference on Computer Vision and
  Pattern Recognition (CVPR)}, pages 304--313, 2019.

\bibitem[Ronneberger et~al.(2015)Ronneberger, Fischer, and
  Brox]{ronneberger2015u}
Olaf Ronneberger, Philipp Fischer, and Thomas Brox.
\newblock {U-Net: Convolutional Networks for Biomedical Image Segmentation}.
\newblock In \emph{International Conference on Medical Image Computing and
  Computer-Assisted Intervention}, pages 234--241. Springer, 2015.

\bibitem[Esser et~al.(2018)Esser, Sutter, and Ommer]{esser2018variational}
Patrick Esser, Ekaterina Sutter, and Bj{\"o}rn Ommer.
\newblock {A variational U-Net for Conditional Appearance and Shape
  Generation}.
\newblock In \emph{Proceedings of the IEEE Conference on Computer Vision and
  Pattern Recognition (CVPR)}, pages 8857--8866, 2018.

\bibitem[Dorkenwald et~al.(2020)Dorkenwald, Buchler, and
  Ommer]{dorkenwald2020unsupervised}
Michael Dorkenwald, Uta Buchler, and Bjorn Ommer.
\newblock {Unsupervised Magnification of Posture Deviations Across Subjects}.
\newblock In \emph{Proceedings of IEEE Conference on Computer Vision and
  Pattern Recognition (CVPR)}, pages 8256--8266, 2020.

\bibitem[Baur et~al.(2018)Baur, Wiestler, Albarqouni, and Navab]{baur2018deep}
Christoph Baur, Benedikt Wiestler, Shadi Albarqouni, and Nassir Navab.
\newblock Deep autoencoding models for unsupervised anomaly segmentation in
  brain mr images.
\newblock In \emph{International MICCAI Brainlesion Workshop}, pages 161--169.
  Springer, 2018.

\bibitem[Kingma and Ba(2014)]{kingma2014adam}
Diederik~P Kingma and Jimmy Ba.
\newblock {Adam: A Method for Stochastic Optimization}.
\newblock \emph{arXiv preprint arXiv:1412.6980}, 2014.

\bibitem[Zolfaghari et~al.(2017)Zolfaghari, Oliveira, Sedaghat, and
  Brox]{zolfaghari2017chained}
Mohammadreza Zolfaghari, Gabriel~L Oliveira, Nima Sedaghat, and Thomas Brox.
\newblock {Chained Multi-stream Networks Exploiting Pose, Motion, and
  Appearance for Action Classification and Detection}.
\newblock In \emph{Proceedings of the IEEE International Conference on Computer
  Vision}, pages 2904--2913, 2017.

\bibitem[Vyas et~al.(2020)Vyas, Rawat, and Shah]{vyas2020multiview}
Shruti Vyas, Yogesh~S Rawat, and Mubarak Shah.
\newblock {Multiview Action Recognition Using Cross-View Video Prediction}.
\newblock In \emph{Proceedings of the European Conference on Computer Vision
  (ECCV)}, 2020.

\bibitem[Shi et~al.(2015)Shi, Chen, Wang, Yeung, Wong, and
  Woo]{shi2015convolutional}
Xingjian Shi, Zhourong Chen, Hao Wang, Dit~Yan Yeung, Wai~Kin Wong, and
  Wang~Chun Woo.
\newblock {Convolutional LSTM Network: A Machine Learning Approach for
  Precipitation Nowcasting}.
\newblock \emph{Proceedings of the Advances in Neural Information Processing
  Systems}, 2015:\penalty0 802--810, 2015.

\bibitem[Misra et~al.(2016)Misra, Zitnick, and Hebert]{misra2016shuffle}
Ishan Misra, Lawrence Zitnick, and Martial Hebert.
\newblock {Shuffle and Learn: Unsupervised Learning using Temporal Order
  Verification}.
\newblock In \emph{European Conference on Computer Vision (ECCV)}, pages
  527--544. Springer, 2016.

\bibitem[Luo et~al.(2017)Luo, Peng, Huang, Alahi, and
  Fei-Fei]{luo2017unsupervised}
Zelun Luo, Boya Peng, De-An Huang, Alexandre Alahi, and Li~Fei-Fei.
\newblock {Unsupervised Learning of Long-term Motion Dynamics for Videos}.
\newblock In \emph{Proceedings of the IEEE Conference on Computer Vision and
  Pattern Recognition (CVPR)}, pages 2203--2212, 2017.

\bibitem[Yao et~al.(2021)Yao, Zhao, Xie, Ye, and Liang]{yao2021recurrent}
Han Yao, SJ~Zhao, Chi Xie, Kenan Ye, and Shuang Liang.
\newblock {Recurrent Graph Convolutional Autoencoder for Unsupervised
  Skeleton-Based Action Recognition}.
\newblock In \emph{2021 IEEE International Conference on Multimedia and Expo
  (ICME)}, pages 1--6. IEEE, 2021.

\bibitem[Su et~al.(2020)Su, Liu, and Shlizerman]{su2020predict}
Kun Su, Xiulong Liu, and Eli Shlizerman.
\newblock {Predict and Cluster: Unsupervised Skeleton based Action
  Recognition}.
\newblock In \emph{Proceedings of the IEEE Conference on Computer Vision and
  Pattern Recognition (CVPR)}, pages 9631--9640, 2020.

\bibitem[Lin et~al.(2020)Lin, Song, Yang, and Liu]{lin2020ms2l}
Lilang Lin, Sijie Song, Wenhan Yang, and Jiaying Liu.
\newblock {MS2L: Multi-Task Self-Supervised Learning for Skeleton Based Action
  Recognition}.
\newblock In \emph{Proceedings of the 28th ACM International Conference on
  Multimedia}, pages 2490--2498, 2020.

\bibitem[Rao et~al.(2021)Rao, Xu, Hu, Cheng, and Hu]{rao2021augmented}
Haocong Rao, Shihao Xu, Xiping Hu, Jun Cheng, and Bin Hu.
\newblock {Augmented Skeleton based Contrastive Action Learning with Momentum
  Lstm for Unsupervised Action Recognition}.
\newblock \emph{Information Sciences}, 569:\penalty0 90--109, 2021.

\bibitem[Kay et~al.(2017)Kay, Carreira, Simonyan, Zhang, Hillier,
  Vijayanarasimhan, Viola, Green, Back, Natsev, et~al.]{kay2017kinetics}
Will Kay, Joao Carreira, Karen Simonyan, Brian Zhang, Chloe Hillier, Sudheendra
  Vijayanarasimhan, Fabio Viola, Tim Green, Trevor Back, Paul Natsev, et~al.
\newblock {The Kinetics Human Action Video Dataset}.
\newblock \emph{arXiv preprint arXiv:1705.06950}, 2017.

\end{thebibliography}
\end{document}